\documentclass[10pt,twocolumn,letterpaper]{article}

\usepackage{epsfig}
\usepackage{amsmath}
\usepackage{amssymb}
\usepackage{hhline}
\usepackage{color, colortbl}
\usepackage{multirow}
\usepackage{multicol}
\usepackage{graphicx}
\usepackage{subcaption}
\usepackage[utf8]{inputenc}
\usepackage{lipsum}  
\usepackage{stfloats}
\usepackage{booktabs}

\usepackage{xcolor}
\usepackage{makecell}
\usepackage{url}

\usepackage{authblk}
\usepackage{multirow}
\usepackage[overload]{empheq}
\usepackage{nopageno}
\usepackage{listings}

\begin{document}

%%%%%%%%% TITLE
\title{Analysis of Adversarial Image Manipulations}

\author{Ahsi Lo}
\author{Gabriella Pangelinan}
\author{Michael C. King}
\affil{Florida Institute of Technology, Melbourne, FL}
\maketitle
%\thispagestyle{empty}

%%%%%%%%% ABSTRACT
\begin{abstract}
As virtual and physical identity grow increasingly intertwined, the importance of privacy and security in the online sphere becomes paramount. In recent years, multiple news stories have emerged of private companies scraping web content and doing research with or selling the data. Images uploaded online can be scraped without users' consent or knowledge. Users of social media platforms whose images are scraped may be at risk of being identified in other uploaded images or in real-world identification situations. This paper investigates how simple, accessible image manipulation techniques affect the accuracy of facial recognition software in identifying an individual's various face images based on one unique image.
\end{abstract}

%%%%%%%%% BODY TEXT
\section{Introduction}
 In recent years, news articles \cite{scrapeimages,carrsmyth_2021,keith_2021} have reported on the problem of facial recognition databases using images scraped from the web without the knowledge or consent of users. In 2021, the United States Government Accountability Office (GAO) noted that the facial recognition databases of 20 federal agencies contain ``millions or billions of photos'' - the Federal Bureau of Investigation (FBI)'s, for example, contains ``over 640 million photos, in some cases [obtained] through the use of private companies that scrape social media sites.'' \cite{keith_2021} While the procurement and usage of the scraped images alone is enough to cause concern to the public, perhaps the most unsettling aspect is the fact that these scraped images may be used by nefarious agents to find other online instances of image subjects. 
 
 How, then, can an individual protect their privacy when uploading images to the web? In this paper, we examine adversarial image manipulation techniques users can apply to their images in an effort to decrease the accuracy of facial recognition systems if their images are scraped. (That is, decrease the likelihood that a scraped image instance may be used as a probe in a search to find other image instances of the subject.) We consider seven techniques which were chosen for their accessibility and ease / speed of application with minimal impact to image style. 
 
 To simulate the scraped-image web-seach situation, we use unique ``probe'' images of 500 individuals and compare the accuracy of facial verification and identification methods using unmanipulated versus manipulated probes. The goal of this project is to determine which manipulations are most effective in decreasing recognition accuracy while (1) being simple for users to implement and (2) maintaining the style of the image.

The three image manipulation techniques we investigate are (1) variation of brightness, (2) application of filters, and (3) addition of a cloak.
% [@Gabby], 
Cloaking is a concept used by the Fawkes algorithm from The SAND Lab at University of Chicago \cite{fawkes}. It makes tiny pixel-level changes to images which are imperceptible to the human eye but impactful to facial recognition models of a person. 

According to a report by the National Institute of Standards and Technology (NIST), variation in illumination is one of the most challenging factors for facial recognition systems to overcome \cite{lightandfocus}. High-brightness regions in images typically correspond to areas of overexposure \cite{hasikin2012enhancement}. To the human eye, overexposed facial regions appear very light or white. Similarly, from a computer's perspective, overexposed image regions will be full of white pixels and can minimize valuable information about facial components and their relationships, thereby decreasing the effectiveness of feature detection and extraction. Alternately, low-brightness facial regions indicate underexposure. Underexposed regions appear dark and shadowy to a human observer and as dark or black pixels to a computer. Like the previous case, underexposure can blur facial landmarks and negatively impact recognition accuracy.

Coloration of an image plays a role in the efficacy of detection and feature extraction of a face image. An image's color distribution can affect ``estimation of the boundaries, shapes, and sizes of facial attributes such as eyes and hairlines'' \cite{colortofacerecog}. Grayscale, for example, affects the contrast (amount of texture) of the image, with amount of texture being an important factor in estimating facial boundaries \cite{grayscale}.

We select this study's manipulation techniques based on research into how lighting and image quality affects facial recognition algorithms. To investigate brightness effects, we adjust images to be either brighter or darker using a Python library. For coloration, we use another Python library to apply ``Instagram-like'' filters on images. Instagram is a social media platform that allows users to upload images and apply readily available filters that change the coloration of their image. Our final manipulation is the Fawkes cloaking algorithm because it is widely available to users of Windows and Mac operating systems through a simple application. 
%-------------------------------------------------------------------------
\section{Background and Related Work}

Previous research on face recognition has focused on three key areas: image-based, structure-based, and function-based approaches. For image-based approaches, researchers study the effects of image-level face information, including edge- and surface-based information. Edge information refers to discontinuities in a facial image and is obtained via contours, features, and other characteristics of the face. Pigmentation, shading, and shadow fall under surface-based information; shading and shadow are a result of lighting \cite{imagebased}. According to Hu et al. \cite{onlineandillum}, severe illumination variations - such as those present in uncontrolled, in-the-wild imagery - can decrease facial recognition accuracy. Since web-scraped images are typically intentionally posted and therefore tend to have more controlled lighting, deep learning methods trained on these images tend to perform poorly on uncontrolled imagery \cite{onlineandillum}. There is also a relationship between how lighting and focus affect face recognition performance. Beverage et al. \cite{lightandfocus} found that edge density, which is impacted by lighting, affects face recognition and high edge density is associated with outdoor images with harsh lighting while low edge density is associated with more uniform lighting. They also found that (1) recognition fails when a face is illuminated from the side by direct sunlight because it causes strong shadows and strong asymmetry and (2) focus does not make recognition easier or harder \cite{lightandfocus}.

Abaza et al. \cite{qualitymeasures} explored how different image quality factors such as contrast, brightness, sharpness, focus, and illumination affect face recognition. Contrast refers to the difference in color intensity that distinguishes objects. Brightness refers to intensity of light emitted and ranges from bright to dim. Focus refers to clarity of detail - that is, degree of blurring or sharpness. Illumination describes the amount of light that passes through or is emitted from an area in the image. When evaluating the effects of these image quality factors, Abaza et al. found that face recognition performance decreases when contrast increases but is generally consistent (with slight changes) across a range of brightness levels. In regards to focus and sharpness, face recognition performance decreases with increased smoothing. Lastly, performance is reasonable for minor illumination changes but degrades dramatically with major illumination changes \cite{qualitymeasures}. Other research on image quality factors yielded similar results. Ke et al. \cite{sharpandbright} found that when changing the luminance of images by brightening and dimming, recognition rate will decrease gradually. Their Receiver Operating Characteristic (ROC) curves for images with different brightness shows that image quality declines with change in brightness. Furthermore, the reduction of clarity caused by blurring an image makes recognition rate decrease significantly \cite{sharpandbright}.

Many of these image quality factors such as contrast, brightness, and focus can be edited using photo editing applications. However, they have not been proven to thwart face recognition algorithms. One image manipulation method that does proclaim achievement of ``100\% success in experiments against today's state-of-the-art facial recognition services'' is Fawkes, a system that was developed by The SAND Lab at University of Chicago \cite{fawkes}. Fawkes uses a cloaking algorithm that creates minimal perturbations in a facial image, altering the feature set of the face with small changes almost invisible to the human eye. Their goal is for users to upload cloaked images of themselves online and unauthorized facial recognition models that collected the users' online images would have a distorted view of the users' feature set and later would be unable to recognize uncloaked images of the user. Using Fawkes, users would add cloaks onto their images before uploading online and the cloaks would not impede normal use of the image. A limitation of Fawkes is that users may not be able to control all images of themselves uploaded online, i.e. if their image is uploaded by friends or family. In this case, Shan et al. recommend proactively monitoring and removing tags of their name (i.e. tagged images featuring them) on social media \cite{fawkes}.

In terms of uploading images onto social media, one very popular social media platform is Instagram. Instagram allows users to upload their media onto their profiles and view media from other users they follow or popular users through the exploration page. When a user uploads an image onto Instagram, they have the option to edit their photo and apply a predetermined Instagram filter. Instagram filters allow modifications of images through selection of pre-defined filters and allow users to apply visual effects to make their images aesthetically appealing. Wu et al. \cite{instagram} studied the effects of 20 popular Instagram filters on convolutional neural network (CNN)-based recognition systems: they found that dramatic changes in appearance as a result of Instagram filters led to differences in feature sets. This, in turn, worsened system performance. In fact, when using pre-trained models ResNet50 and IBN with original images versus Instagram-filtered images, they noted that rank-1 accuracy dropped by 8.92\% (ResNet50) and 7.89\% (IBN). This may be due to filters injecting additional style information into feature maps, which changes the feature representations. The authors proposed a de-stylization module which ``removes'' style information from feature maps and re-normalizes the maps to transform filtered-image features to approximate those of the original image. The de-stylization module predicts the parameters used to normalize feature maps and uses this information to remove style information encoded in a filtered image; the module was found to increase rank-1 accuracy for their training dataset \cite{instagram}.

Overall, the goal of this study is to explore the effects of changing brightness, coloration, and cloaking on facial recognition using a freely available online face dataset, public Python libraries, and the Fawkes application which is available on the Fawkes website. Our choices of dataset and methods make our experiments easily replicable by other researchers.
%-------------------------------------------------------------------------
\section{Dataset and Preprocessing}
We use the Celebrities in Frontal-Profile in the Wild (CFPW) dataset \cite{cfp-paper} for this study. CFPW contains 500 subjects, each with ten frontal and four profile images. 

The data preprocessing workflow differs slightly for the verification versus identification experiments. Both workflows consider the same set of seven image manipulations, given in Table \ref{tab:transformationTable}.

\begin{table}[!ht]
\centering
\resizebox{\columnwidth}{!}{%
\begin{tabular}{|c|l|l|l|}
\hline
\textbf{Subject}     & \multicolumn{1}{c|}{\textbf{Manipulation}} & \multicolumn{1}{c|}{\textbf{Probe Name}} & \multicolumn{1}{c|}{\textbf{Probe Set}} \\ \hline
\multirow{7}{*}{001} & Increase Brightness                          & ``001\_probe\_bright.jpg"                 & Probes\_Bright                          \\ \cline{2-4} 
                     & Decrease Brightness                          & ``001\_probe\_dark.jpg"                   & Probes\_Dark                            \\ \cline{2-4} 
                     & Apply Cloak                                  & ``001\_probe\_cloak.jpg"                  & Probes\_Cloak                           \\ \cline{2-4} 
                     & Apply Filter 1                               & ``001\_probe\_f1.jpg"                     & Probes\_Filter1                         \\ \cline{2-4} 
                     & Apply Filter 2                               & ``001\_probe\_f2.jpg"                     & Probes\_Filter2                         \\ \cline{2-4} 
                     & Apply Filter 3                               & ``001\_probe\_f3.jpg"                     & Probes\_Filter3                         \\ \cline{2-4} 
                     & Apply Filter 4                               & ``001\_probe\_f4.jpg"                     & Probes\_Filter4                         \\ \hline
\end{tabular}%
}
\caption{Manipulations applied  to  each  subject’s  probe image, shown for Subject 1.}
\label{tab:transformationTable}
\end{table}

\subsection{Verification Task Preprocessing}
\begin{figure}[!h]
    \centering
    \includegraphics[width=\columnwidth]{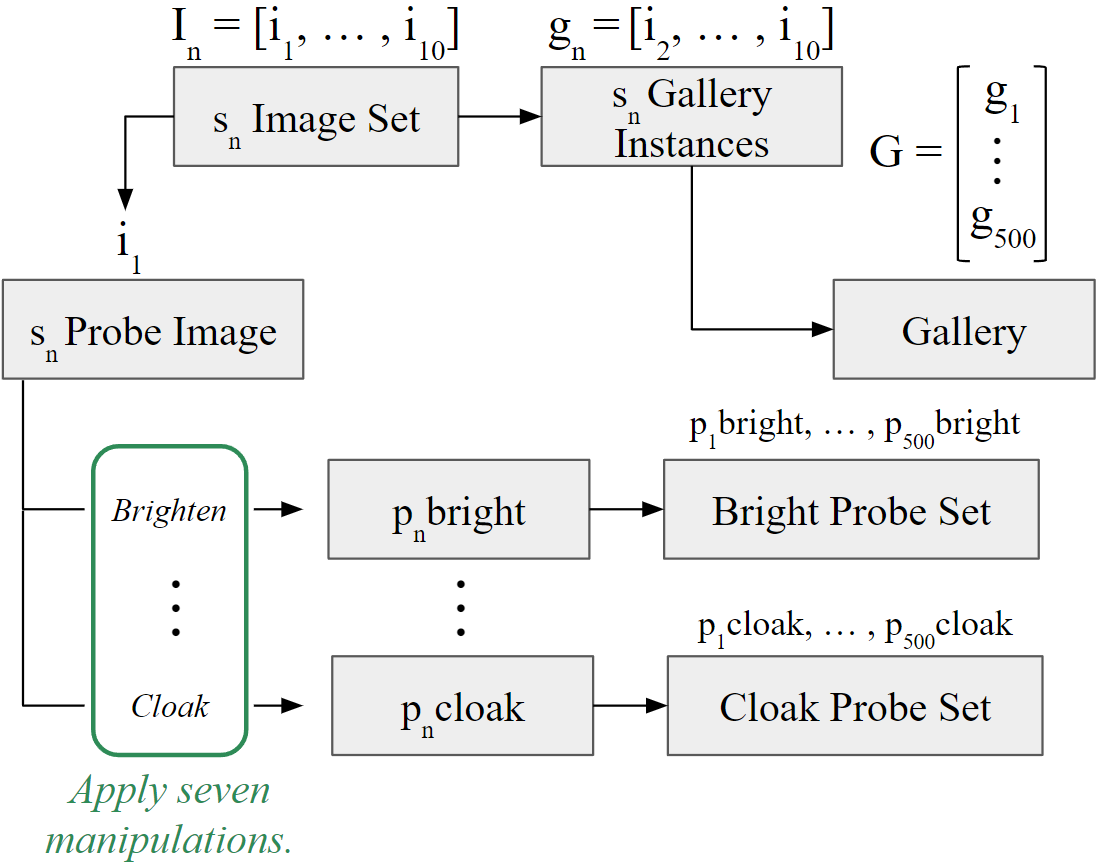}\hfill
    \caption{Verification task: data preprocessing workflow.}
    \label{fig:expMethod_Verif}
\end{figure}

For each subject $\text{s}_n$ in the subject set S = $\text{s}_1, ... , \text{s}_{500}$, the verification task preprocessing steps (visualized in Figure \ref{fig:expMethod_Verif}) are as follows:
\begin{enumerate}
    \item From $\text{s}_n$'s image set I = $\text{i}_1, ... , \text{i}_{10}$, randomly select one image $\text{i}_n$ as $\text{s}_n$'s probe. Without loss of generality, assume probe is $\text{i}_1$.
    \item Append $\text{s}_n$'s unmanipulated images $\text{i}_2, ... , \text{i}_{10} \in I$ to the gallery.
    \item Append the nine non-probe instances of all other subjects ($s_2,...,s_{500})$ to the gallery.
    \item Create seven copies of $\text{i}_1$.
    \item Apply one manipulation (given in Table \ref{tab:transformationTable}) to each copy of $\text{i}_1$.
    \item Append each manipulated probe-copy to its corresponding probe set.
\end{enumerate}

\subsection{Identification Task Preprocessing}
The identification task preprocessing workflow differs only in Step 2 (namely, the number of probe subject images to add to the gallery). Steps 1 and 2 for identification preprocessing are given below; steps 3-6 are the same.
\begin{enumerate}
    \item From $\text{s}_n$'s image set I = $\text{i}_1, ... , \text{i}_{10}$, randomly select one image $\text{i}_n$ as $\text{s}_n$'s probe. Without loss of generality, assume probe is $\text{i}_1$.
    \item Randomly select a single image $\text{i}_n (\ne \text{i}_1)$ from $\text{s}_n$'s image set I (suppose $\text{i}_{10}$) and append it to the gallery.
\end{enumerate}

\begin{figure}[!h]
    \centering
    \includegraphics[width=\columnwidth]{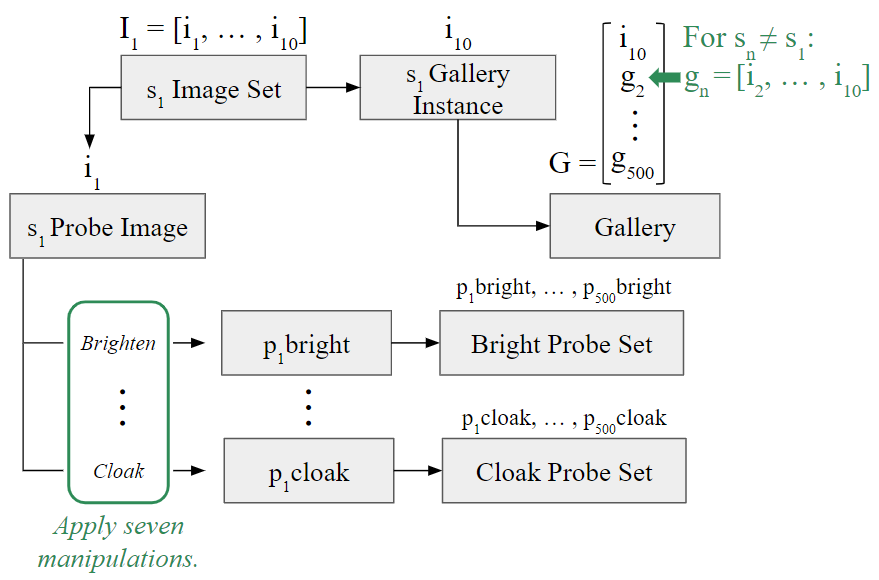}\hfill
    \caption{Identification task: data preprocessing workflow.}
    \label{fig:expMethod_Ident}
\end{figure}

Figure \ref{fig:expMethod_Ident} gives the identification-specific workflow.

\subsection{Image Manipulations}
Seven manipulations, given in Table \ref{tab:transformationTable}, were applied to images. The first two manipulations require adjusting probe brightness. We used the Image Enhance function of the \emph{Python Imaging Library (PIL)} \cite{pil} to brighten and darken the image.

\begin{figure}[h!]
     \centering
     \includegraphics[width=\columnwidth]{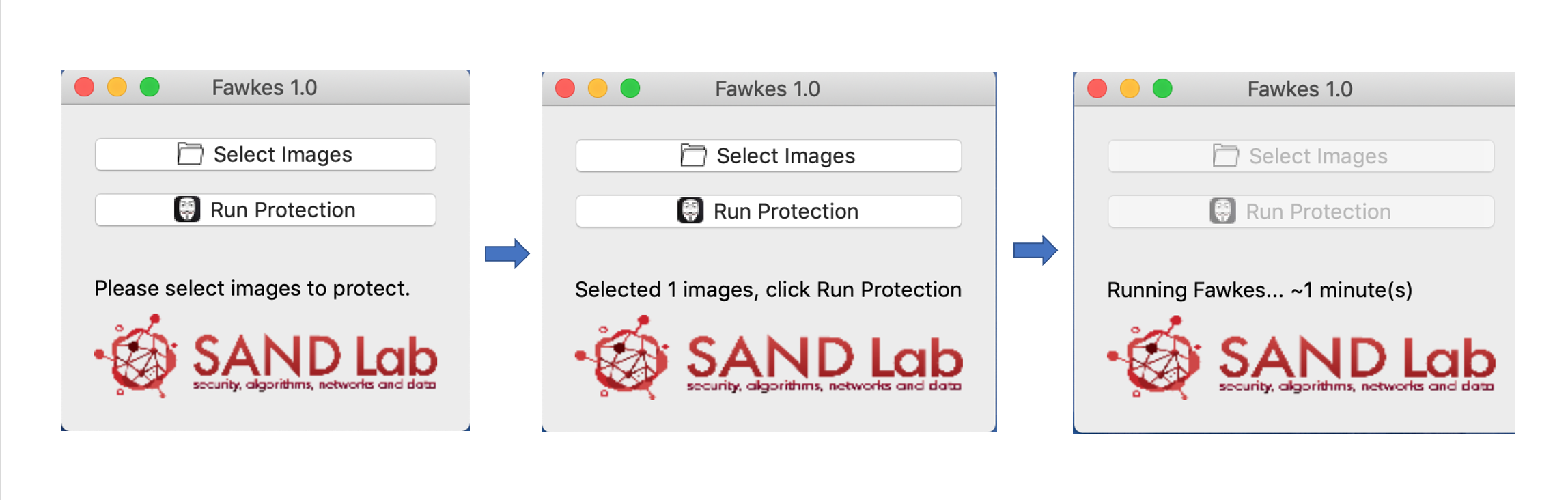}\hfill
     \caption{Fawkes cloaking GUI.}
     \label{fig:fawkesGUI}
 \end{figure}

For the ``cloaking'' manipulation, we used The SAND Lab's Fawkes application (v1.0). The application takes approximately 60 seconds to process an image and return a cloaked version. While the Fawkes app's graphical user interface (GUI) (shown in Figure \ref{fig:fawkesGUI}) simplifies the addition of cloaks, its runtime degrades with larger image sets. Since the app does not indicate time remaining until completion, it may seem unresponsive when processing a large number of images.

For coloration, the Python library \emph{pilgram} \cite{pilgram} was used. Four Instagram filters (``Lofi'', ``Moon'', ``X-Pro II'', and ``Earlybird'') were selected based on visual inspection with the intention of maximally varying image coloration while retaining image integrity. 
\begin{itemize}
    \item X-Pro II adds a vignette effect (darkened edges and corners) and intensifies shadows \cite{instagramfil1}.
    \item Lo-Fi adds shadows and increases color saturation.
    \item Earlybird gives a yellow tint and adds a vignette and soft blurring effect \cite{earlybird}.
    \item Moon grayscales the image and intensifies shadows.
\end{itemize}

\begin{figure}[!h]
    \centering
    \includegraphics[width=.8\columnwidth]{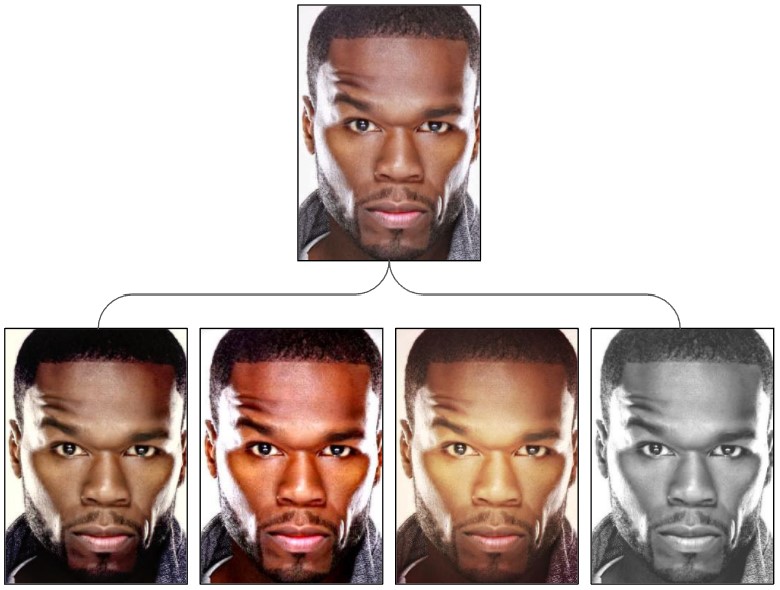}\hfill
    \caption{Original image (top) and its four filtered versions (from left to right: X-Pro II, Lo-Fi, Earlybird, Moon).}
    \label{fig:compFilters}
\end{figure}

The application of each filter to a sample subject probe is shown in Figure \ref{fig:compFilters}.

%-------------------------------------------------------------------------
\section{Experimental Results}
In this experiment, we generate match scores between probe and gallery images using Rank One, a commercial off-the-shelf (COTS) facial recognition software. In the Department of Homeland Security's 2020 Biometric Technology Rally \cite{rally} for assessment of biometric algorithm capabilities, Rank One ``achieved at least 99\% accuracy with 3 of the 6 different acquisition systems.'' \cite{roc} For a given probe image, Rank One's verification function is used to obtain similarity scores between the probe and each gallery image. If similarity is detected between a probe and gallery image, a match score between 0 and 1 is assigned, with higher values indicating greater perceived similarity. If no similarity is detected, a match score of 0 is assigned to the pair.

We can view the recognition of manipulated images as both a verification (1:1) and identification (1:N) problem. As a verification problem, we assess whether a face in an unmanipulated gallery image of a subject is correctly verified (that is, has a similarity score above a given threshold) as the probe image face. As an identification problem, we consider the rank-1 identification rate for each probe set and the maximum rank required to return a single gallery instance of the probe subject.

The average genuine and impostor scores for each manipulated set, with percent change in scores versus the original (unmanipulated) probes, are given in Tables \ref{tab:avgGens} and \ref{tab:avgImps}.
\begin{table}[!h]
\centering
\resizebox{.8\columnwidth}{!}{%
\begin{tabular}{|c|c|c|}
\hline
\textbf{\begin{tabular}[c]{@{}c@{}}Image\\ Manipulation\end{tabular}} & \textbf{\begin{tabular}[c]{@{}c@{}}Average \\ Genuine Score\end{tabular}} & \textbf{\begin{tabular}[c]{@{}c@{}}\% Change \\ vs. Baseline\end{tabular}} \\ \hline
\cellcolor[HTML]{D9D9D9}None                                          & \cellcolor[HTML]{D9D9D9}0.97589                                            & -                                                                          \\ \hline
Bright                                                                & 0.96692                                                                    & -0.92\%                                                                    \\ \hline
Dark                                                                  & 0.97320                                                                    & -0.28\%                                                                    \\ \hline
XPro                                                                  & 0.97177                                                                    & -0.42\%                                                                    \\ \hline
LoFi                                                                  & 0.97209                                                                    & -0.39\%                                                                    \\ \hline
Earlybird                                                             & 0.96787                                                                    & -0.82\%                                                                    \\ \hline
Moon                                                                  & 0.97167                                                                    & -0.43\%                                                                    \\ \hline
Cloak                                                                 & 0.94921                                                                    & -2.73\%                                                                    \\ \hline
\end{tabular}%
}
\caption{Average genuine match scores for each manipulation.}
\label{tab:avgGens}
\end{table}
\begin{table}[!h]
\centering
\resizebox{.8\columnwidth}{!}{%
\begin{tabular}{|c|c|c|}
\hline
\textbf{\begin{tabular}[c]{@{}c@{}}Image\\ Manipulation\end{tabular}} & \textbf{\begin{tabular}[c]{@{}c@{}}Average \\ Impostor Score\end{tabular}} & \textbf{\begin{tabular}[c]{@{}c@{}}\% Change \\ vs. Baseline\end{tabular}} \\ \hline
\cellcolor[HTML]{D9D9D9}None                                          & \cellcolor[HTML]{D9D9D9}0.06824                                            & -                                                                          \\ \hline
Bright                                                                & 0.06793                                                                    & -0.45\%                                                                    \\ \hline
Dark                                                                  & 0.06784                                                                    & -0.59\%                                                                    \\ \hline
XPro                                                                  & 0.06803                                                                    & -0.31\%                                                                    \\ \hline
LoFi                                                                  & 0.06841                                                                    & +0.25\%                                                                    \\ \hline
Earlybird                                                             & 0.06777                                                                    & -0.69\%                                                                    \\ \hline
Moon                                                                  & 0.06802                                                                    & -0.32\%                                                                    \\ \hline
Cloak                                                                 & 0.06852                                                                    & +0.41\%                                                                    \\ \hline
\end{tabular}%
}
\caption{Average impostor match scores for each manipulation.}
\label{tab:avgImps}
\end{table}

\subsection{Verification Analysis}
Here, we analyze the False Reject Rate (FRR) and ROC curve for manipulated probe sets against the gallery. Table \ref{tab:frrTable} gives the FRR for each manipulation with thresholds of 0.8 (column 2) and 0.9 (column 4). Columns 3 and 5 give the percent increase of each manipulation's FRR from the baseline FRR (highlighted in gray). For both thresholds, addition of a cloak is the most impactful manipulation in increasing FRR, followed by increased image brightness and application of the Earlybird filter. At a  threshold of 0.9, cloak yielded a +163.3\% increase in FRR followed by Increased Brightness at +54.6\% change and Earlybird filter at +46.9\% change. The lowest FRR change was LoFi at +19.6\% change. At a threshold of 0.9, the ranking of most impactful to least impactful image manipulations was Cloak, Increased Brightness, Earlybird, Moon, XPro, Decreased Brightness, and LoFi. Interestingly, at a threshold of 0.8, the top three manipulations with highest FRR-change remained the same but the lowest-change manipulations were XPro (+34.5\% increase), Moon (+31.4\% increase), LoFi (+27.5\% increase), and finally Decreased Brightness (+14.8\% increase). 
\begin{table}[!ht]
\centering
\resizebox{\columnwidth}{!}{%
\begin{tabular}{|c|c|c|c|c|}
\hline
\textbf{\begin{tabular}[c]{@{}c@{}}Image\\ Manipulation\end{tabular}} & \textbf{\begin{tabular}[c]{@{}c@{}}FRR1 \\ t = 0.8\end{tabular}} & \textbf{\begin{tabular}[c]{@{}c@{}}FRR1 \\ \% Change\end{tabular}} & \textbf{\begin{tabular}[c]{@{}c@{}}FRR2 \\ t = 0.9\end{tabular}} & \textbf{\begin{tabular}[c]{@{}c@{}}FRR2 \\ \% Change\end{tabular}} \\ \hline
\cellcolor[HTML]{D9D9D9}None (Baseline)                               & \cellcolor[HTML]{D9D9D9}0.0357\%                                 & \textbf{-}                                                         & \cellcolor[HTML]{D9D9D9}0.0815\%                                 & \textbf{-}                                                         \\ \hline
\rowcolor[HTML]{FFFFFF} 
\cellcolor[HTML]{FFFFFF}Bright                                        & \cellcolor[HTML]{FFFFFF}0.061\%                                  & +70.9\%                                                            & 0.126\%                                                          & +54.6\%                                                            \\ \hline
\rowcolor[HTML]{FFFFFF} 
\cellcolor[HTML]{FFFFFF}Dark                                          & \cellcolor[HTML]{FFFFFF}0.041\%                                  & +14.8\%                                                            & 0.098\%                                                          & +20.2\%                                                            \\ \hline
\rowcolor[HTML]{FFFFFF} 
\cellcolor[HTML]{FFFFFF}XPro                                          & \cellcolor[HTML]{FFFFFF}0.048\%                                  & +34.5\%                                                            & 0.099\%                                                          & +21.5\%                                                            \\ \hline
\rowcolor[HTML]{FFFFFF} 
\cellcolor[HTML]{FFFFFF}LoFi                                          & \cellcolor[HTML]{FFFFFF}0.0455\%                                 & +27.5\%                                                            & 0.0975\%                                                         & +19.6\%                                                            \\ \hline
\rowcolor[HTML]{FFFFFF} 
\cellcolor[HTML]{FFFFFF}Earlybird                                     & \cellcolor[HTML]{FFFFFF}0.0577\%                                 & +61.6\%                                                            & 0.1197\%                                                         & +46.9\%                                                            \\ \hline
\rowcolor[HTML]{FFFFFF} 
\cellcolor[HTML]{FFFFFF}Moon                                          & \cellcolor[HTML]{FFFFFF}0.0469\%                                 & +31.4\%                                                            & 0.1024\%                                                         & +25.6\%                                                            \\ \hline
\rowcolor[HTML]{FFFFFF} 
\cellcolor[HTML]{FFFFFF}Cloak                                         & \cellcolor[HTML]{FFFFFF}0.1086\%                                 & +204.2\%                                                           & 0.2146\%                                                         & +163.3\%                                                           \\ \hline
\end{tabular}%
}
\caption{FRR of each manipulation and comparison to baseline FRR, using two thresholds.}
\label{tab:frrTable}
\end{table}

FRR is a meaningful metric because in the real-world image-scraping setting, we want to decrease the number of ``true'' image instances of an individual that can be found online using a single image (probe) of the individual. So, we can relate an increased FRR to increased likelihood that a true instance of the individual will \emph{not} be returned by a web search (i.e. a true instance will be ``falsely rejected''). 
\begin{figure}[!h]
    \centering
    \includegraphics[width=\columnwidth]{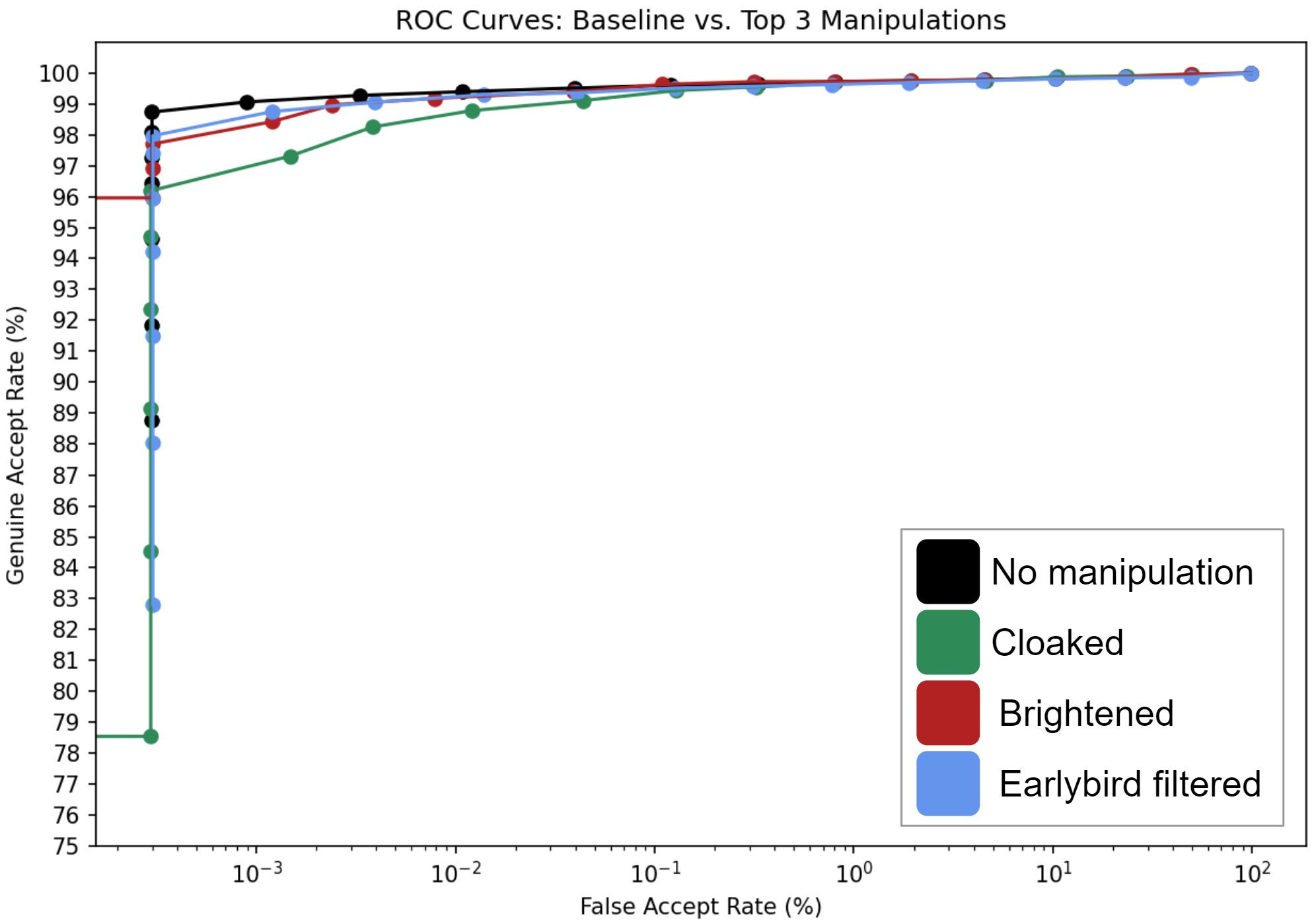}\hfill
    \caption{Comparison of ROC curves for unmanipulated images versus the top three manipulations from Table \ref{tab:frrTable}.}
    \label{fig:rocComparison}
\end{figure}

Figure \ref{fig:rocComparison} shows the ROC curves for the baseline case (probe images with no manipulation) versus the three manipulations with the greatest FRR change: cloaking, brightening, and Earlybird-filtering. (In accordance with standard convention, FAR values (x-axis) are given in log-scale.) 

Using \emph{PyEER} \cite{pyeer}, a Python package for performance evaluation of biometric systems, we calculate the system's equal error rate (EER) for each manipulation type (shown in Table \ref{tab:eerTable}). EER is the threshold at which FAR and FRR are equal, and a lower EER value indicates better system performance. Interestingly, system EER decreased (albeit very minimally) for four manipulations (darkened, Moon- and Earlybird-filtered, and brightened) and increased (minimally) for three (cloaking, XPro- and LoFi-filtered).
\begin{table}[!h]
\centering
\resizebox{.5\columnwidth}{!}{%
\begin{tabular}{|c|c|}
\hline
\textbf{\begin{tabular}[c]{@{}c@{}}Image\\ Manipulation\end{tabular}} & \textbf{\begin{tabular}[c]{@{}c@{}}Equal Error \\ Rate (\%)\end{tabular}} \\ \hline
\rowcolor[HTML]{D9D9D9} 
None (Baseline)                                                       & 99.3268\%                                                                 \\ \hline
Bright                                                                & 99.3228\%                                                                 \\ \hline
Dark                                                                  & 99.3182\%                                                                 \\ \hline
XPro                                                                  & 99.3395\%                                                                 \\ \hline
LoFi                                                                  & 99.3403\%                                                                 \\ \hline
Earlybird                                                             & 99.3204\%                                                                 \\ \hline
Moon                                                                  & 99.3197\%                                                                 \\ \hline
Cloak                                                                 & 99.337\%                                                                  \\ \hline
\end{tabular}%
}
\caption{System EER for each manipulation's probe set.}
\label{tab:eerTable}
\end{table}

\subsection{Identification Analysis}
In the identification experiment, for each probe subject, only one gallery instance of the subject was present. Table \ref{tab:rankTable} gives the rank-1 identification rate and maximum rank for each manipulation. We define ``max rank returned'' (column 4) as the maximum rank required to correctly identify all subjects.

\begin{table}[!ht]
\centering
\resizebox{\columnwidth}{!}{%
\begin{tabular}{|c|c|c|c|c|}
\hline
\textbf{\begin{tabular}[c]{@{}c@{}}Image\\ Manipulation\end{tabular}} & \textbf{\begin{tabular}[c]{@{}c@{}}Rank-1 \\ Ident. \\ Rate (\%)\end{tabular}} & \textbf{\begin{tabular}[c]{@{}c@{}}\% Change \\ vs. Baseline\end{tabular}} & \textbf{\begin{tabular}[c]{@{}c@{}}Max \\ Rank\end{tabular}} & \textbf{\begin{tabular}[c]{@{}c@{}}\% Change \\ vs. Baseline\end{tabular}} \\ \hline
\cellcolor[HTML]{D9D9D9}None                                          & \cellcolor[HTML]{D9D9D9}99.6\%                                                 & -                                                                          & \cellcolor[HTML]{D9D9D9}42                                   & -                                                                          \\ \hline
\cellcolor[HTML]{FFFFFF}Bright                                        & 99.6\%                                                                         & -                                                                          & 59                                                           & +40.48\%                                                                   \\ \hline
\cellcolor[HTML]{FFFFFF}Dark                                          & 99.6\%                                                                         & -                                                                          & 54                                                           & +28.57\%                                                                   \\ \hline
\cellcolor[HTML]{FFFFFF}XPro                                          & 99.6\%                                                                         & -                                                                          & 108                                                          & +157.14\%                                                                  \\ \hline
\cellcolor[HTML]{FFFFFF}LoFi                                          & 99.6\%                                                                         & -                                                                          & 37                                                           & -11.9\%                                                                    \\ \hline
\cellcolor[HTML]{FFFFFF}Earlybird                                     & 99.6\%                                                                         & -                                                                          & 31                                                           & -26.19\%                                                                   \\ \hline
\cellcolor[HTML]{FFFFFF}Moon                                          & 99.6\%                                                                         & -                                                                          & 79                                                           & +88.1\%                                                                    \\ \hline
\cellcolor[HTML]{FFFFFF}Cloak                                         & 99.8\%                                                                         & +0.2\%                                                                     & 134                                                          & +219.05\%                                                                  \\ \hline
\end{tabular}%
}
\caption{Rank-1 identification rate and maximum rank of returned instances for each manipulation.}
\label{tab:rankTable}
\end{table}

There was little change in the rank-1 identification rate across image manipulations; only the cloaked probes showed a 0.2\% increase. Maximum rank, however, increased for five of the seven manipulations and decreased for two. The manipulations with highest max rank were Cloak (+219.05\% increase), followed by XPro (+157.14\% increase), and Moon (+88.1\% increase) and with the lowest were Earlybird (-26.19\% decrease) and LoFi (-11.9\% decrease).

In future work, we will consider datasets with demographic-specific groupings. Performing these experiments on larger image sets of individuals who share more apparent facial qualities by nature of race and gender (e.g. global face shape, skin tone, softness of features) may provide more insight into identification accuracy given image manipulations.
%-------------------------------------------------------------------------
\section{Conclusion and Discussion}
This study investigates the effect of various image manipulations on facial recognition accuracy. We consider these manipulations adversarial in that they are intended to decrease facial recognition accuracy. 

As a verification problem, we found that FRR increased for each manipulation versus the baseline. EER decreased for four manipulations and increased for three, indicating that some manipulations do have the desired negative effect on system performance. In comparing the effects of brightness, coloration, and cloaking, we saw that the manipulations that yielded the highest FRR were cloaking, increased brightness, and the Earlybird Instagram filter. These results are consistent with previous research done on face recognition. In our experiment, we found that cloaking does provide a good means of decreasing the accuracy of facial recognition, although results are comparative to use of other filters and were not significant. Furthermore, the fact that the currently available Fawkes cloaking GUI is slow and can appear unresponsive on larger image sets may not make it an ideal solution if high-volume processing is desired. Increased brightness, as it corresponds to overexposure, minimizes valuable facial information that can be used to describe facial components and the relationships between them. Thus, it is reasonable that increased brightness would yield a higher FRR. Lastly, the Earlybird filter added style elements to the original image through yellow tint, vignette, and soft blur. This is consistent with previous research that increased smoothing decreases facial recognition performance. 

Four manipulations (decreasing brightness and application of Moon, XPro and LoFi filters) had 25.6\% or less change in FRR versus the baseline. The likely reason these particular manipulations had the least impact on facial recognition performance was that the intensification of shadows or color saturation may not be as intense as other changes, like the smoothing and blurring of Earlybird. 

As an identification problem, we found that rank-1 identification rate was unchanging for nearly all manipulations (with a very minimal decrease for the cloaked probe images). Maximum rank required to return all probe set subjects' single gallery instances increased for five of seven manipulations and decreased for two. In the future, we will consider a much larger number of subjects and use demographic groupings to yield increased similarities in facial attributes between subjects and determine whether rank-1 accuracy decreases.

In future work, we will merge different image manipulation techniques - specifically, the combination of cloaking with either brightness or filter adjustments - and determine their combined impact on face recognition. The coloration-cloaking combination would represent a real world situation of an individual taking a facial image of themselves, applying the Fawkes cloaking, and choosing an Instagram filter before uploading online. If this combination of manipulation yields a higher distance score, that would support the use of the Fawkes algorithm for posting images on social media platforms like Instagram. Also, this could relate to previous work on de-stylizing methods to ``undo'' applied filters and ``revert'' a filtered image to the original, as we did not find much existing research in reversing the effects of cloaking combined with Instagram filters. We will also consider expanding on and adjusting other image quality factors such as focus and sharpness, contrast, and illumination to test what level of each factor is still acceptable to the human eye and which factors, in isolation or combined, have the greatest effect of face recognition. Another area relating to filter adjustments is experimenting with all of the available Instagram filters to determine which filter causes the greatest impact on face recognition.

Ultimately, though the task of undermining facial recognition systems can be somewhat challenging given the robustness and ever-improving nature of modern systems, this study has shown that increasing user privacy and peace-of-mind via minimal image manipulations is a worthwhile endeavor with promising results.
%-------------------------------------------------------------------------
%{\small
%\bibliographystyle{ieee}
%\bibliography{egbib.bib}
%}

%\end{document}

{\small
\bibliographystyle{ieee_fullname}
\bibliography{egbib}

\begin{thebibliography}{10}\itemsep=-1pt

\bibitem{rally}
Matching system results for the 2020 biometric technology rally, Jan 2021.

\bibitem{roc}
Rank one continues strong performance in most recent dhs biometric technology
  rally, Mar 2021.

\bibitem{qualitymeasures}
A. Abaza, M. Harrison, T. Bourlai, and A Ross.
\newblock Design and evaluation of photometric image quality measures for
  effective face recognition.
\newblock In {\em IET Biometrics}, pages 314--324, June 2014.

\bibitem{lightandfocus}
J.~R. Beveridge, D.~S. Bolme, B.~A. Draper, G.~H. Givens, Y.~M. Lui, and P.~J.
  Phillips.
\newblock Quantifying how lighting and focus affect face recognition
  performance.
\newblock In {\em 2010 IEEE Computer Society Conference on Computer Vision and
  Pattern Recognition - Workshops}, pages 74--81, 2010.

\bibitem{carrsmyth_2021}
Julie Carr~Smyth.
\newblock States push back against use of facial recognition by police.
\newblock In {\em The Associated Press}, May 2021.

\bibitem{instagramfil1}
M. Gola.
\newblock 7 best instagram filters to make it more effective.
\newblock In {\em Curvearro}, September 2020.

\bibitem{hasikin2012enhancement}
Khairunnisa Hasikin and Nor Ashidi~Mat Isa.
\newblock Enhancement of the low contrast image using fuzzy set theory.
\newblock In {\em 2012 UKSim 14th International Conference on Computer
  Modelling and Simulation}, pages 371--376. IEEE, 2012.

\bibitem{scrapeimages}
Hill.
\newblock The secretive company that might end privacy as we know it.
\newblock In {\em The New York Times}, January 2020.

\bibitem{onlineandillum}
C. Hu, F. Wu, J. Yu, X. Jing, X. Lu, and P. Liu.
\newblock Diagonal symmetric pattern-based illumination invariant measure for
  severe illumination variation face recognition.
\newblock In {\em IEEE Access}, pages 63202--63213, 2020.

\bibitem{pilgram}
Akiomi Kamakura.
\newblock pilgram.
\newblock \url{https://github.com/akiomik/pilgram}, 2020.

\bibitem{sharpandbright}
L. Ke, H. Chen, F. Huang, S. Ling, and Z. You.
\newblock Sharpness and brightness quality assessment of face images for
  recognition.
\newblock In {\em Scientific Programming}, pages 1--21, 2021.

\bibitem{keith_2021}
Morgan Keith.
\newblock 20 federal agencies use facial recognition technologies that store
  billions of photos.
\newblock In {\em Yahoo! News}, Jul 2021.

\bibitem{grayscale}
Michael King.
\newblock Introduction to face recognition (ppt).
\newblock Florida Institute of Technology, 2021.

\bibitem{pyeer}
Manuel Martinez.
\newblock Pyeer.
\newblock \url{https://github.com/manuelaguadomtz/pyeer}, 2021.

\bibitem{earlybird}
M Mayne.
\newblock How to make instagram filters in photoshop: Earlybird.
\newblock In {\em Photodoto}, n.d.

\bibitem{cfp-paper}
S. Sengupta, J.C. Cheng, C.D. Castillo, V.M. Patel, R. Chellappa, and D.W.
  Jacobs.
\newblock Frontal to profile face verification in the wild.
\newblock In {\em IEEE Conference on Applications of Computer Vision}, February
  2016.

\bibitem{fawkes}
S. Shan, E. Wenger, J. Zhang, H. Li, H. Zheng, and B.~Y. Zhao.
\newblock Fawkes: Protecting personal privacy against unauthorized deep
  learning models.
\newblock In {\em USENIX Security Symposium 2020}, 2020.

\bibitem{pil}
Hugo van Kemenade.
\newblock Python imaging library (pil).
\newblock \url{https://github.com/python-pillow/Pillow}, 2021.

\bibitem{instagram}
Zhe Wu, Zuxuan Wu, Bharat Singh, and Larry~S. Davis.
\newblock Recognizing instagram filtered images with feature de-stylization.
\newblock 2019.

\bibitem{imagebased}
Jisien Yang and Shyi~Gary C.W.
\newblock Image-based approach to face recognition: Effects of line-drawn
  faces, pigmentation and shading, and spatial frequency.
\newblock In {\em Chinese Journal of Psychology}, pages 99--127, 2002.

\bibitem{colortofacerecog}
A. Yip and P. Sinha.
\newblock Contribution of color to face recognition.
\newblock In {\em Perception}, pages 995--1003, 2002.

\end{thebibliography}
}

\end{document}